\documentclass[10pt,twocolumn,letterpaper]{article}

\usepackage{cvpr}              

\usepackage{graphicx}
\usepackage{amsmath}
\usepackage{amssymb}
\usepackage{booktabs}
\usepackage{xcolor}
\usepackage{subcaption}
\usepackage{multicol}
\usepackage{float}
\usepackage{fancyhdr}

\usepackage[capitalize]{cleveref}
\crefname{section}{Sec.}{Secs.}
\Crefname{section}{Section}{Sections}
\Crefname{table}{Table}{Tables}
\crefname{table}{Tab.}{Tabs.}

\def\cvprPaperID{*****} 
\def\confName{OAGM}
\def\confYear{2022}

\begin{document}

\title{Dimensionality reduction for AI based hyperspectral image classification based on XAI}

\author{Vladimir Zeljković\\
University of Belgrade School of Electrical Engineering\\
JOANNEUM RESEARCH Forschungsgesellschaft mbH\\
{\tt\small zv200450d@student.etf.bg.ac.rs}
\and
Branka Stojanović\\
JOANNEUM RESEARCH Forschungsgesellschaft mbH\\
{\tt\small Branka.Stojanovic@joanneum.at}
\and
Harald Ganster\\
JOANNEUM RESEARCH Forschungsgesellschaft mbH\\
{\tt\small Harald.Ganster@joanneum.at}
\and
Aleksandar Nešković\\
University of Belgrade School of Electrical Engineering\\
{\tt\small neshko@etf.rs}
}
\maketitle

\begin{abstract}
This research addresses the challenge of limited material recycling in wood recycling processes by leveraging artificial intelligence (AI)-based dimensionality reduction. Our study explores the application of convolutional neural networks (CNNs) in multi-channel hyperspectral imaging (HSI), extending beyond RGB channels to over 200 spectral channels. Dimensionality reduction within this context involves streamlining the feature space for AI system training and inference. Focusing on explainable AI (XAI) methods, this paper contributes to a broader research initiative, presenting a solution framework that enhances the sustainability and efficiency of wood recycling processes.
\end{abstract}

\section{Introduction}\label{int}

The research described in this paper focuses on the problem of limited material recycling in wood recycling processes. One major hurdle is our inability to easily recognize and separate different materials, especially when trying to automate the process. This issue becomes even more critical in today's world, where products are designed to be used for shorter periods. To tackle this problem, we aim to find ways to make recycling more cost-effective and eco-friendly by efficiently reusing limited raw materials with minimal energy and processing resources. 

This study, as part of a large research initiative \cite{Stojanovic2023}, utilizes AI for different materials classification based on hyperspectral imaging (HSI) and explores in detail a specific aspect of this problem -- dimensionality reduction, to make wood recycling more sustainable and efficient. AI methods selected for the experiments include the most commonly used deep learning approach for image processing -- convolutional neural networks (CNNs), and extend those methods from 3-channel space (RGB channels for visible light images) to multi-channel space, with more than 200 channels. These channels include sampled wavelength in a certain range. 
Dimensionality reduction in this context refers to reducing a feature space (number of channels) used for AI system training and later inference.


This paper is organized in the following way: Section 2 describes the hyperspectral imaging  problem space and related work. Section 3 describes the proposed solution and selected XAI method for the experiments. Section 4 describes results and their implications. The conclusion finally discusses the findings and future work on this project.

\section{Problem space}\label{prob}

Hyperspectral imaging founds its extensive application in remote sensing and medical domains. Nevertheless, different application areas like material characterisation became more prominent in the last couple of years. This section describes the problem space, including the brief technological background and related work in this domain. 

\subsection{Hyperspectral imaging in material characterisation}

The aim of our research initiative is to develop future-oriented methods for the compositional characterisation of complex material streams. A case study used in this paper is the wood recycling process. Wood recycling material streams can have a wide range of compositions, from pre-sorted recyclables to unsorted residual waste, including items of different sizes. The different types of materials typically include raw wood, processed wood (e.g. MDF, HDF...) and various contaminants that need to be removed prior to the recycling process, such as plastics, glass, paper, textiles, inert materials (e.g. stones, bricks...).
A data acquisition technology foreseen in our study is the use of a conveyor belt in combination with a Short Wave Infrared (SWIR) line scanner covering the spectrum from 960nm to 2550nm, sampled in 288 discrete channels. The general idea is to implement artificial intelligence methods to identify the composition of these material streams based on the spectral data. Given that the spectrum contains a large number of channels (288 in our case), dimensionality reduction by pre-selecting the wavelengths of interest is envisaged as a method that could bring numerous benefits. On the one hand, it could have a significant impact on the processing speed and performance of the system, reducing computational overhead. This, in turn, leads to more real-time or efficient processing, making the system more practical for various applications. On the other hand, dimensionality reduction could significantly reduce the cost of processing hyperspectral data, and it also extends the lifetime of the acquisition equipment, which can become strained by the large volume of data. In essence, reducing the number of wavelengths not only enhances system efficiency but also makes the implementation more cost-effective and sustainable in the long term.

\subsection{Related work}

Extensive literature analysis reveals two primary AI-based classification approaches using hyperspectral imaging (HSI) data: (i) object detection and classification, leveraging advanced object detectors like YOLOv3 \cite{Pang2022} and YOLOv5 \cite{Qingyun2021}, which excel in surveillance and specific object detection scenarios; and (ii) pixel-based region classification \cite{Ahmad2021}, offering versatility by not relying on specific object characteristics, making it better suited for diverse material characterization tasks, especially when objects vary in shape and overlap on conveyor belts.

When it comes to dimensionality reduction in the HSI domain, the most commonly used method is principal components analysis (PCA), e.g. in \cite{Pang2022}. In addition to PCA method, there are different methods proposed in the literature, including enhanced hybrid-graph discriminant learning method \cite{Luo2020}, local linear embedding \cite{Fang2014}, spectral segmentation \cite{Siddiqa2022}, etc. Study \cite{Moharram2023} proposes an extensive survey of dimensionality reduction methods in the remote sensing area.

\section{Dimensionality reduction based on XAI}\label{method}

\subsection{Proposed solution}

To initiate our preliminary assessments, we opted for Convolutional Neural Networks (CNNs), a prevalent machine learning and deep learning technique extensively employed in image classification tasks within the literature \cite{Ahmad2021}. In particular, we evaluated two distinct CNN architectures: Conv1D and Conv2D \cite{Ahmad2021, Stojanovic2023}, both adept at handling multichannel data inputs, in contrast to conventional image processing algorithms that typically operate on images with three channels (RGB). Our study \cite{Stojanovic2023}, includes an extensive testing of these two methods and explains the detailed configuration. 

In this study we use HSI data cubes of the fixed width (384 pixels) for the training and testing of proposed methods, captured with the SWIR visual sensor, covering wavelengths in the SWIR range, sampled into the 288 discreet channels. We used a subset of HSI cubes for training, validation and initial testing, containing different types of materials of interest for our use case. These HSI cubes were partitioned into non-overlapping regions (patches), subsequently allocated to training, validation and testing sets. A series of tests were conducted in our previous study \cite{Stojanovic2023} to ascertain the optimal patch size from the HSI cube, encompassing dimensions of 3×3, 5×5, 7×7, 9×9, and 11×11. Based on this work we have selected a 5x5 patch size for the experiments described in this paper. Another subset of HSI cubes containing samples of different materials were used for testing inference and results visualisation.

The experiments described in this paper utilize Conv1D network for dimensionality reduction based on XAI, and Conv2D for material classification. Different material classes in our experiment include: raw wood, type 1 processed wood, type 2 processed wood, type 3 processed wood, glass, inert, plastic and paper.

\subsection{XAI dimensionality reduction}
In this section, we delve into the dimensionality reduction approach employed in our research, specifically focusing on the use of Grad-CAM (Gradient-weighted Class Activation Mapping). Grad-CAM, a popular XAI technique, plays a pivotal role in uncovering the importance of individual wavelengths within hyperspectral data for image classification.

\subsubsection{Grad-CAM: The Chosen XAI Method}

The idea for using Grad-CAM as our XAI method for hyperspectral images was inspired by the research described in the paper
\cite{DeLucia2022}. This source highlighted its effectiveness in enhancing the interpretability of AI models for hyperspectral image analysis.

Grad-CAM stands as a prominent choice among XAI techniques for its ability to offer insights into the decision-making process of CNNs. It excels in highlighting which regions or wavelengths in hyperspectral images contribute significantly to the network's classification decisions. The selection of Grad-CAM aligns with its capacity to enhance the interpretability of AI models, a critical factor in hyperspectral image analysis. This method empowers us to gain profound insights into the importance of individual wavelengths in making classification decisions.


In our research, we employ a specialized Conv1D designed to process hyperspectral data effectively, which serves as the backbone for Grad-CAM analysis. Our network consists of a single convolutional layer with a kernel size of 5, signifying that each computed value in the network is influenced by five consecutive wavelengths. This architectural choice allows us to capture local wavelength features effectively.

\subsubsection{Grad-CAM: How it Works} \label{GCAM}

Grad-CAM operates through a series of steps that unravel the importance of individual wavelengths in hyperspectral material classification. In this section, we delve into the inner workings of Grad-CAM, offering insights into the mathematical foundations and its role in our research.

\paragraph{Gradient Computation}

Grad-CAM starts by calculating the gradients of the predicted class score concerning the feature maps from the final convolutional layer of the CNN. These gradients quantify how much each feature map contributed to the classification decision.

Mathematically, the gradient computation can be represented as follows:

\[
\frac{\partial C}{\partial A_k} \frac{\partial A_k}{\partial \text{ReLU}} = \frac{\partial C}{\partial \text{ReLU}}
\]

Where:

\begin{itemize}
\item \(\frac{\partial C}{\partial A_k}\) represents the gradient of the predicted class score \(C\) with respect to the feature map activation \(A_k\) from the final convolutional layer.
\end{itemize}

\paragraph{Global Average Pooling (GAP)}

After gradient computation, we introduce a unique step: Global Average Pooling. In our case, the gradient feature map's size is equal to the number of convolutional layer filters (20) multiplied by the number of wavelengths after convolution.

GAP calculates the average gradient value for each filter, effectively reducing the spatial dimensions while retaining the information about the importance of specific wavelengths.

Mathematically, GAP can be represented as averaging the gradients across all wavelengths for each of the filters:

\[
\text{GAP}(x)_c = \frac{1}{W} \sum_{i=1}^{W} x_c(i)
\]

Where:

\begin{itemize}
\item \(\text{GAP}(x)_c\) represents the channel-wise average gradient value for the \(c\)-th filter.
\item \(W\) stands for the number of wavelengths after convolution.
\end{itemize}

\paragraph{Filter-wise Weighting}

The average gradient values derived from GAP serve as importance scores for each filter within the feature maps. This is a pivotal step where we emphasize the contributions of more relevant wavelengths to the classification process. It essentially assigns weights to each channel based on its influence on the classification outcome.

Grad-CAM's core operation involves performing weighted combinations of feature maps. The mean gradient values obtained through GAP act as weights, scaling each feature map. After this weighting process, we obtain a heatmap that highlights the importance of different wavelengths for the final classification decision. This indexed heatmap provides valuable insights into which wavelengths and regions of the hyperspectral data are significant for the network's classification process.


\subsubsection{Input Data}

The input data consists of 64 samples for each class material, resulting in a comprehensive dataset that represents various materials present in hyperspectral images. Each sample is a 5×5 spatial patch of the hyperspectral image.

\paragraph{Data Reshaping}

Before feeding the data into our Conv1D network, we perform data reshaping. Each 5×5 spatial patch contains 288 wavelength channels. To make this data suitable for the 1D CNN, we reshape it into a $25 \times (number of wavelengths)$ format. This reshaping ensures that our CNN can effectively process the data while considering the spectral information from the hyperspectral image.

\paragraph{Class-level Heatmap}

For each sample in the dataset, we calculate a heatmap using the Grad-CAM technique, as discussed in \cref{GCAM}, and based on that
we compute a class-level heatmap. This heatmap is obtained by taking the mean value of the individual heatmaps calculated for the 64 samples within a class. The class-level heatmap represents the collective importance of wavelengths for a specific class material, represented by the importance index in the range $(0,1)$.

This data preprocessing and heatmap generation process contribute significantly to our dimensionality reduction approach using Grad-CAM, allowing us to gain a deeper understanding of the hyperspectral data's spectral features and their impact on material classification.

\subsubsection{Selection of Important Wavelengths from Heatmap}

\begin{figure*}[h]
  \centering
  \begin{subfigure}{0.33\linewidth}
    \includegraphics[width=\linewidth]{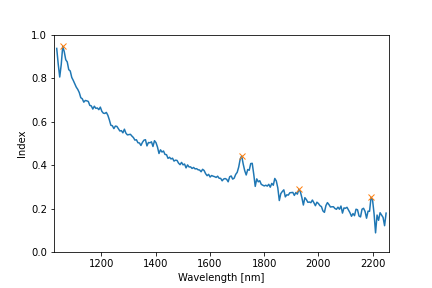}
    \caption{Background}    
  \end{subfigure}
  \hfill
  \begin{subfigure}{0.33\linewidth}
    \includegraphics[width=\linewidth]{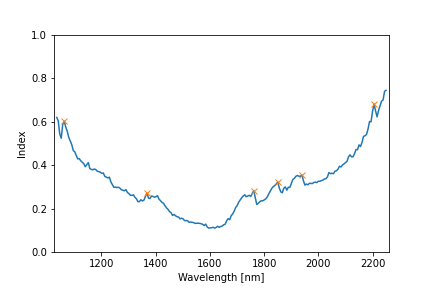}
    \caption{Glass}
  \end{subfigure}
  \hfill
  \begin{subfigure}{0.33\linewidth}
    \includegraphics[width=\linewidth]{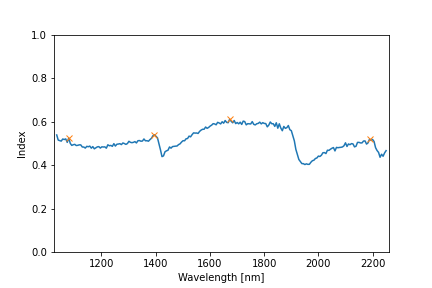}
    \caption{Inert}
  \end{subfigure}

  \begin{subfigure}{0.33\linewidth}
    \includegraphics[width=\linewidth]{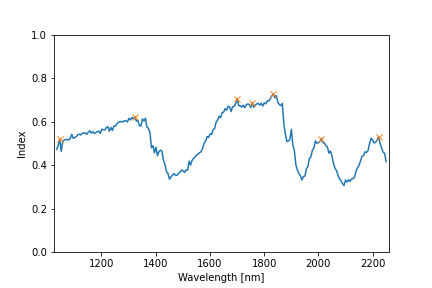}
    \caption{Paper}
  \end{subfigure}
  \hfill
  \begin{subfigure}{0.33\linewidth}
    \includegraphics[width=\linewidth]{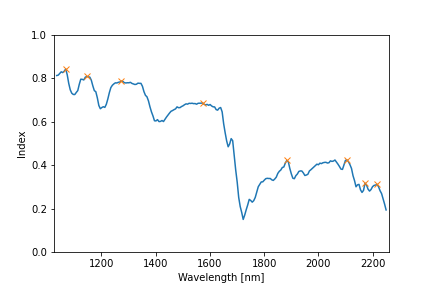}
    \caption{Plastic}
  \end{subfigure}
  \hfill
  \begin{subfigure}{0.33\linewidth}
    \includegraphics[width=\linewidth]{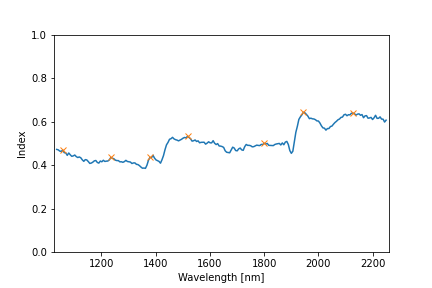}
    \caption{Type 1 Processed Wood}
  \end{subfigure}

  \begin{subfigure}{0.33\linewidth}
    \includegraphics[width=\linewidth]{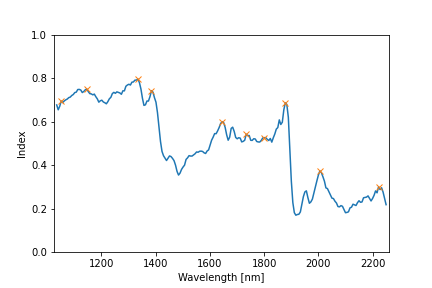}
    \caption{Type 2 Processed Wood}
  \end{subfigure}
  \hfill
  \begin{subfigure}{0.33\linewidth}
    \includegraphics[width=\linewidth]{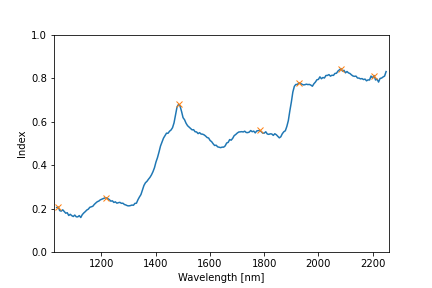}
    \caption{Type 3 Processed Wood}
  \end{subfigure}
  \hfill
  \begin{subfigure}{0.33\linewidth}
    \includegraphics[width=\linewidth]{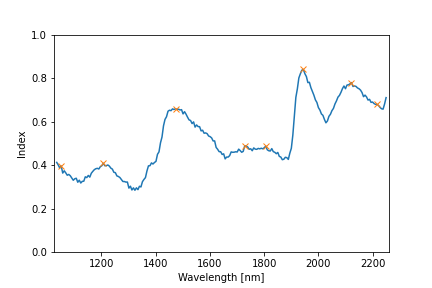}
    \caption{Wood}
  \end{subfigure}
  
  \caption{Averaged heatmaps for different data classes.}
  \label{fig:heatmaps}
  
\end{figure*}

\cref{fig:heatmaps} contains images that were generated using the process described in the previous section. Each image represents an averaged heatmap for a specific class. To identify important wavelengths, we employed a method that involves identifying local maxima within these heatmap functions. 
Selected important wavelengths for each class are marked with $x$ on the graphics.
\\
\\
The selected maxima were then sorted based on their respective values in the graphics, allowing us to discern the most significant wavelengths for each class.

\section{Evaluation and Discussion}\label{eval}
In this section, we evaluate the effectiveness of our dimensionality reduction approach using Grad-CAM in hyperspectral material classification. We aim to understand how reducing the number of wavelengths influences classification accuracy. Our assessment involves a series of experiments employing a completely different network architecture -- a Conv2D network. These experiments utilize different sets of wavelengths, varying from single wavelengths identified as significant by Grad-CAM to the entire set of available wavelengths.

\subsection{Performance Evaluation}


Our experiments utilized three distinct sets of wavelengths:

\textbf{Single Wavelength}: This experiment employs only the most crucial wavelength identified by Grad-CAM for each material class.

\textbf{Top 4 Wavelengths}: For this experiment, we select the four most important wavelengths for each material class, determined through Grad-CAM analysis.

\textbf{All Wavelengths}: As a benchmark, we train the CNN using the complete set of available wavelengths in the hyperspectral data.

\subsection{Visual Results}

\begin{figure*}[ht!]
  \centering
  \begin{subfigure}{0.23\linewidth}
    \includegraphics[width=\linewidth]{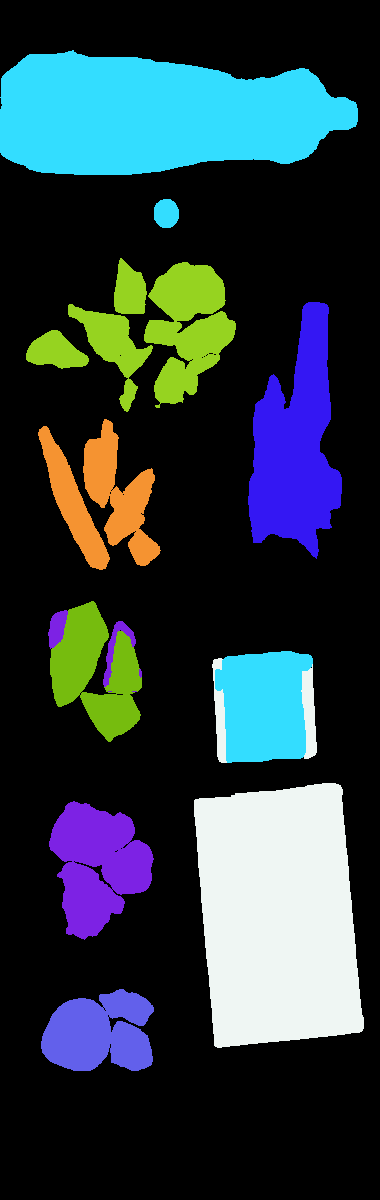}
    \caption{Ground Truth}    
  \end{subfigure}
  \hfill
  \begin{subfigure}{0.23\linewidth}
    \includegraphics[width=\linewidth]{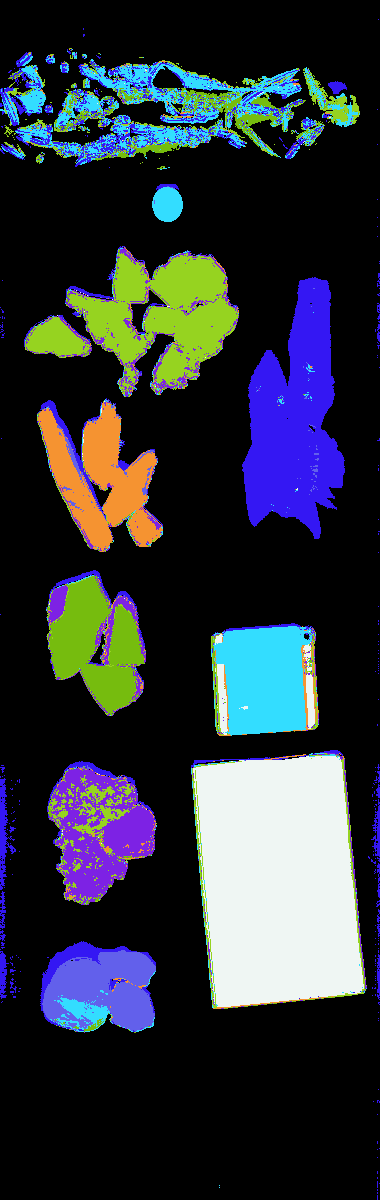}
    \caption{Single Wavelength}
  \end{subfigure}
  \hfill
  \begin{subfigure}{0.23\linewidth}
    \includegraphics[width=\linewidth]{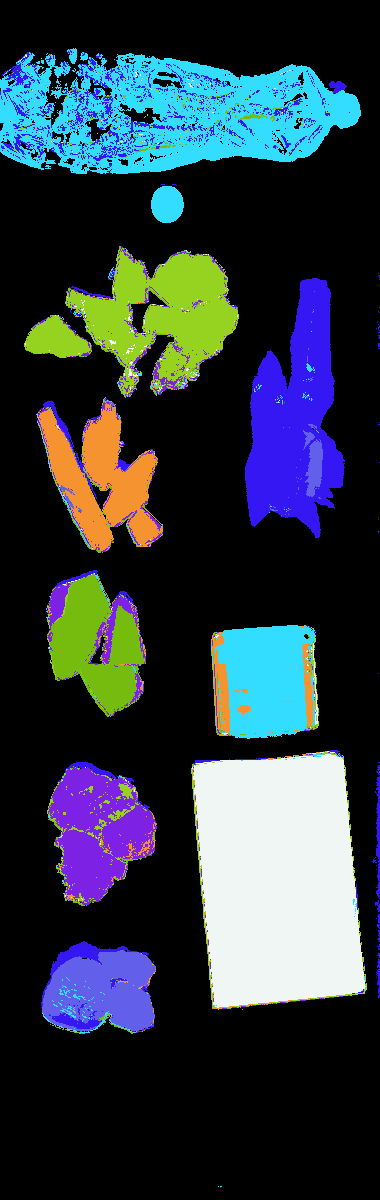}
    \caption{Top 4 Wavelengths}
  \end{subfigure}
  \hfill
  \begin{subfigure}{0.23\linewidth}
    \includegraphics[width=\linewidth]{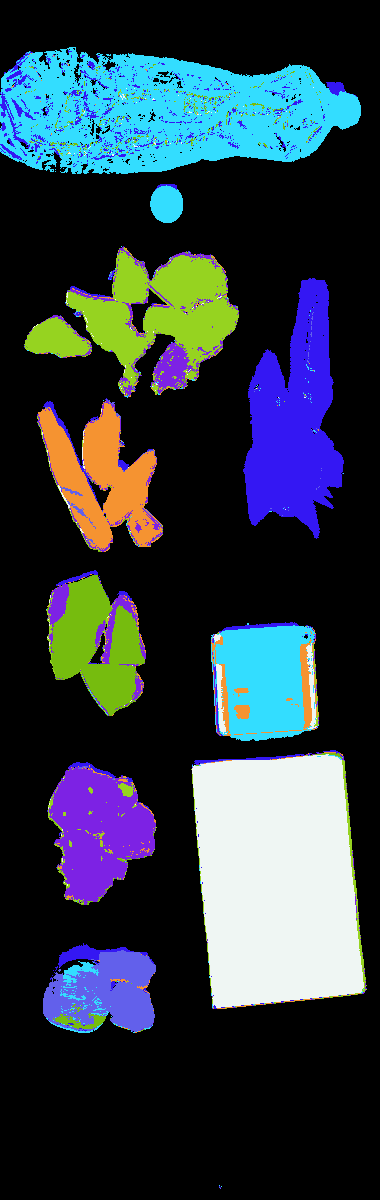}
    \caption{All Wavelengths}
  \end{subfigure}
  \caption{Visual Comparison of Material Classification Results}
  \label{fig:visual_results}
\end{figure*}

To provide a visual understanding of our findings, we present four images that illustrate the classification results for a specific material class using the different sets of wavelengths mentioned above. \cref{fig:visual_results} image includes:
(a) The ground truth label for the material class.
(b) The classification result using only the single most important wavelength for every class.
(c) The classification result using the top four important wavelengths for every class.
(d) The classification result using all available wavelengths.

These images offer a visual comparison of how dimensionality reduction impacts the classification accuracy and showcase the importance of Grad-CAM's role in identifying key wavelengths for accurate material classification.

It can be visually verified that proposed method helps in pre-selecting the important wavelengths and reducing the feature space in a significant matter. 


\subsection{Quantitative Results}

In addition to the visual comparisons, we present quantitative results to provide a more comprehensive assessment of our dimensionality reduction approach. We measured the accuracy of material classification for each experiment, comparing the performance of the CNN using different sets of wavelengths (\cref{tab:visual_results}).

\begin{table*}
\centering
\begin{tabular}{{lccc}}
\hline
\textbf{Class} & \textbf{One Important Wavelength} & \textbf{Four Important Wavelengths} & \textbf{All Wavelengths} \\ \hline
Paper & 98.35\% & 99.84\% & 99.28\% \\ \hline
Plastic & 38.94\% & 74.49\% & 82.71\% \\ \hline
Wood & 94.58\% & 96.10\% & 95.47\% \\ \hline
Background & 96.60\% & 96.77\% & 96.94\% \\ \hline
Glass & 97.74\% & 91.99\% & 97.35\% \\ \hline
Inert & 84.83\% & 89.14\% & 80.24\% \\ \hline
Type 1 Processed Wood & 78.27\% & 91.34\% & 95.65\% \\ \hline
Type 2 Processed Wood & 92.15\% & 94.00\% & 88.04\% \\ \hline
Type 3 Processed Wood & 95.44\% & 95.43\% & 95.25\% \\ \hline
\end{tabular}
\caption{Accuracy of Material Classification Using Different Wavelength Sets}
\label{tab:visual_results}
\end{table*}

These quantitative results offer insights into the impact of dimensionality reduction on classification accuracy. We aim to demonstrate the feasibility of utilizing a reduced set of important wavelengths while maintaining competitive classification performance.

\subsection{Discussion and Insights}

Here, we discuss the key findings and offer insights into the implications of our experiments.

\paragraph{The Power of Single Wavelengths}

Our experiments revealed that a single important wavelength, identified by Grad-CAM, can be remarkably effective for certain material classes. This is evident in the high accuracy achieved for materials like Paper, where a single wavelength proved to be indicative of the material's presence. This finding underscores the potential for resource-efficient hyperspectral imaging, where the acquisition of a limited set of wavelengths may suffice for accurate material classification.

\paragraph{Achieving Balance with Top 4 Wavelengths}

Expanding the wavelength selection to the top four important wavelengths per class demonstrated a balanced improvement in accuracy across all material classes. This suggests that a modest increase in the number of wavelengths, guided by Grad-CAM, can substantially contribute to improving classification accuracy. For materials with complex spectral signatures, such as Plastic, this approach yielded promising results. However, it's important to note that while the accuracy for Plastic improved significantly, we observed a slight decrease in accuracy for Paper. This indicates that the inclusion of additional wavelengths may introduce some confusion for the model, potentially harming the classification performance of specific materials, especially when working with a limited number of wavelengths. 

\paragraph{The Benchmark of All Wavelengths}

Our experiments using all available wavelengths in the hyperspectral data provide the benchmark for material classification accuracy. While this comprehensive approach achieved the highest accuracy for most material classes, it comes at the cost of increased data acquisition and processing requirements. However, for critical applications where accuracy is paramount, using all wavelengths remains the safest option.

\section{Conclusion}\label{conc}

The main goal of our project is to harness intelligent sensor technologies and data analytics to develop a universally applicable solution for characterizing recyclable materials throughout the entire product lifecycle.

The research results proposed in this paper demonstrate the potential for dimensionality reduction in hyperspectral material classification. By leveraging Grad-CAM to identify key wavelengths, we show that it is possible to achieve high accuracy with a limited subset of wavelengths, reducing the data acquisition burden. These findings have implications for resource-constrained environments and applications where efficient material classification is essential.

Our foreseen future work will tend to expand this research to more material classes, and different use-cases, as well as to develop and test additional dimensionality reduction methods.





\end{document}